\documentclass{trbunofficial}
\usepackage{booktabs}
\begin{document}


\title{Classifying Urban Locations by Cyclist Facility Usage and Directional Alignment: A Validated Camera-Based Framework from Multi-Site Observations}

\TRBauthor*{Ao He}{Ph.D. Student, M.E. Rinker, Sr. School of Construction Management, University of Florida. ORCID: 0009-0005-2107-0088}{aohe@ufl.edu}
\TRBauthor{Mohit Gupta}{Ph.D., Senior Machine Learning Engineer, Doxel.ai}{mgupta70@asu.edu}
\TRBauthor{Deborah Salon}{Associate Professor, School of Geographical Sciences and Urban Planning, Arizona State University. ORCID: 0000-0002-2240-8408}{Deborah.Salon@asu.edu}
\TRBauthor{Ricardo Eiris}{Assistant Professor, School of Sustainable Engineering and the Built Environment, Arizona State University. ORCID: 0000-0001-6904-5352}{reiris@asu.edu}
\TRBauthor{Vivian W.H. Wong}{Assistant Professor, M.E. Rinker, Sr. School of Construction Management, University of Florida. ORCID: 0000-0001-9723-8741}{vivian.wong@ufl.edu}

\AuthorHeaders{He, Gupta, Salon, Eiris, and Wong}
\maketitle


\section{Abstract}
\hfill\break%
\noindent\textbf{Objectives:}~``Wrong side of road'' was the primary crash factor in 13.6 percent of California's fatal and serious-injury bicycle crashes in 2023. Agencies spend heavily on bicycle facilities but lack tools to check whether cyclists use them as designed. We ask whether usage patterns follow facility configurations, and whether higher bike lane usage goes with less wrong-way riding.

\hfill\break%
\noindent\textbf{Methods:}~Motion-triggered cameras covered 19 deployments at 17 locations in Tempe, Arizona, in 2023. Manual review identified 1,963 cycling events; excluding 20 crossings left 1,943 labeled events as the ground-truth analysis set. An automated pipeline on the same captures counts 92.9 percent of the full manual total (per-deployment $r$ = 0.99), and every automated wrong-way call was visually verified. Deployments were classified by threshold rules on facility usage shares.

\hfill\break%
\noindent\textbf{Findings:}~The deployments fall into three types: bike-lane dominant (9), sidewalk dominant (6), and mixed (4). Usage follows the facilities present, but presence does not guarantee use: every sidewalk-dominant location has a bike lane that cyclists bypass. Cyclists in bike lanes rode against the flow in 4.6 percent of events; on sidewalks, which carry two-way traffic, riders opposed the near-side flow in 49.6 percent. Across deployments, bike lane share and wrong-way rate correlate negatively (Spearman $\rho = -0.87$, $p < 0.001$, $n = 16$). An independent volunteer count in the same city shows the same sidewalk-direction pattern ($\rho = 0.79$).

\hfill\break%
\noindent\textbf{Novelty:}~Prior studies measure facility use and travel direction at one or two locations, from continuous video. This study measures both across a multi-site still-camera deployment with a fully hand-verified benchmark, and reframes wrong-way riding as a facility-use problem concentrated on sidewalks.

\hfill\break%
\noindent\textbf{Practical Applications:}~The workflow gives planners a low-cost screen for locations where cycling has shifted to the sidewalk and directional conflict follows, before crash records reveal it.
\newpage


\section{Introduction}

Federal spending on bicycle and pedestrian infrastructure now averages \$1.44 billion a year under the Bipartisan Infrastructure Law, nearly double the level before 2021 \citep{FHWA2022Funding}. California alone has committed about \$930 million over four years, including 265 miles of new and improved bike lanes \citep{Caltrans2024}. Part of what this spending is meant to buy is safety. In California, ``wrong side of road'' was recorded as the primary crash factor in 13.6 percent of fatal and serious-injury bicycle crashes in 2023, the fifth most common factor \citep{SafeTREC2025}. The danger is easy to picture: a cyclist riding against traffic approaches from a direction motorists are not checking \citep{Hunter1996}. State agencies are moving on the same concern. Florida's Pedestrian and Bicycle Strategic Safety Plan targets the elimination of fatal and serious-injury crashes for people walking and biking \citep{FDOT2021}; Washington's statewide Active Transportation Plan assesses where its network falls short of safe walking and cycling \citep{WSDOT2021}; and Arizona's Active Transportation Safety Action Plan sets statewide safety targets and calls for better data on how these facilities are actually used \citep{AZDOT2024}.

After bicycle and pedestrian facilities are built, routine evaluation is still required to record what exists, track usage, and plan maintenance. Counts of how many cyclists pass a location are the common metric, typically collected by human counters over short observation windows or by permanent counting stations \citep{FHWA2022}. This metric does not capture how cyclists spread across the available facilities, or whether they travel in the intended direction. The difference between built and used can be large. In Changsha, China, a video study found that even where bike lanes were available, 44 percent of cyclists rode in the travel lane and another 20 percent used the sidewalk \citep{Gao2020}. Studies in American cities have documented wrong-way riding on one-way facilities as well \citep{Dhakal2018}. Inventories and counts cannot see either failure, and decisions based on them inherit the blind spot.

One way to make these measures more reliable is camera observation. Digitized observations can capture what counts and inventories miss: where each cyclist is positioned on the street and which way they travel \citep{Silva2023, Lowry2024, NabaviNiaki2018}. One side of facility use is spatial: where cyclists travel relative to the bike lane, sidewalk, and travel lane. The other is directional: whether they travel with or against the intended traffic direction. These answer different questions. This study measures both from time-stamped camera observations, validates every measurement against a full manual count, and shows that the two sides are linked: wrong-way riding concentrates where cycling happens on sidewalks.

We turn these two sides into two testable hypotheses. The first is that facility configuration shapes usage patterns: cyclists concentrate in the facilities present, so configuration acts as a necessary but not sufficient condition for how a location is used. The second is that higher bike lane use goes with less wrong-way riding. We observed cyclists with motion-triggered cameras at 19 camera deployments covering 17 locations in Tempe, Arizona, manually reviewed every capture (1,963 cycling events, of which 1,943 non-crossing events form the analysis set), and validated an automated pipeline against that benchmark.

This study makes three contributions. First, we develop a four-stage workflow that constructs cycling events from motion-triggered captures, infers facility use and travel direction, and validates itself against a hand-counted ground truth: its event count is 92.9 percent of the full manual count, its per-deployment counts closely track the manual counts (Pearson $r = 0.99$), and every automated wrong-way call was visually verified. Second, we classify the locations into three recurring types: bike-lane dominant, sidewalk dominant, and mixed. Third, we show that travel direction follows facility use. Wrong-way riding is rare in bike lanes (4.6 percent of bike lane events) and common on sidewalks (49.6 percent), which carry two-way traffic, and the association between bike lane share and wrong-way rate across deployments is strong (Spearman $\rho = -0.87$, $p < 0.001$). An independent volunteer count in the same city reproduces the pattern \citep{TBAG2024}.

The sections that follow review prior work, describe the data and the workflow, present the validation, and report what the locations show about each hypothesis.

\section{Background and Related Work}

Fixed cameras and low-cost sensors have been widely used to monitor non-motorized traffic in urban areas. \citet{Lemarchand2025} deployed motion-triggered field cameras at 25 locations in Tempe, Arizona, to conduct 24-hour counts of daily bicycle, scooter, and skateboard traffic; the present study builds on the same camera network but measures behavior rather than volume. The core focus of such studies is how many people are cycling, rather than where or how they ride. They provide temporal and spatial aggregate data, but do not record which facility cyclists use or which direction they travel. More broadly, non-motorized traffic data collection remains well behind motorized traffic in coverage and consistency \citep{FHWA2022, Allen2021}.

Another line of research uses video observation to record cyclist behavior directly at specific locations. \citet{Silva2023} compared cyclist interactions on two types of facilities, on-street bike lanes and sidewalk bike lanes, through video recorded at four street segments in Munich. \citet{Lowry2024} deployed commercial vision sensors on a redesigned street in Boise, Idaho, and tracked how cyclists and pedestrians used a new protected bike lane, including the share riding against its design direction. Computer vision methods have also been applied to detect near-miss events from video streams \citep{Ibrahim2021, Ibrahim2024NearMiss}, to detect close passes from on-bike video \citep{Li2025}, and to estimate wrong-way cycling ratios from sparsely sampled CCTV video \citep{WWC2024}. These studies show that cameras can capture both facility use and travel direction. But they rely on continuous video, and each covers one street or a handful of segments, so the measures cannot be compared across a larger set of locations.

Other studies have used trajectory data to identify movement patterns and wrong-way riding. \citet{NabaviNiaki2018} used video trajectory clustering at four Montreal locations to analyze cyclist movement at infrastructure discontinuities. \citet{Dhakal2018} analyzed 12,202 GPS tracks from the CyclePhilly app to examine wrong-way riding on one-way streets in Philadelphia, finding that 42 percent of trips included wrong-way segments. Trajectory clustering focuses on movement shapes rather than which facilities get used. GPS tracks record the route but not the cyclist's position within the street, so they cannot separate the bike lane from the sidewalk beside it.

In summary, counting studies work across many locations but stop at volumes. Video observation captures facility use and travel direction, but at one or two locations at a time, and requires continuous video. Trajectory studies link behavior to infrastructure, but describe movement rather than facility use. No existing approach measures facility use and travel direction together across a multi-site deployment of low-power still cameras, with the measurements validated against a full manual count.

\section{Data and Study Design}

\begin{figure}[!htbp]
    \centering
    \includegraphics[width=0.8\linewidth]{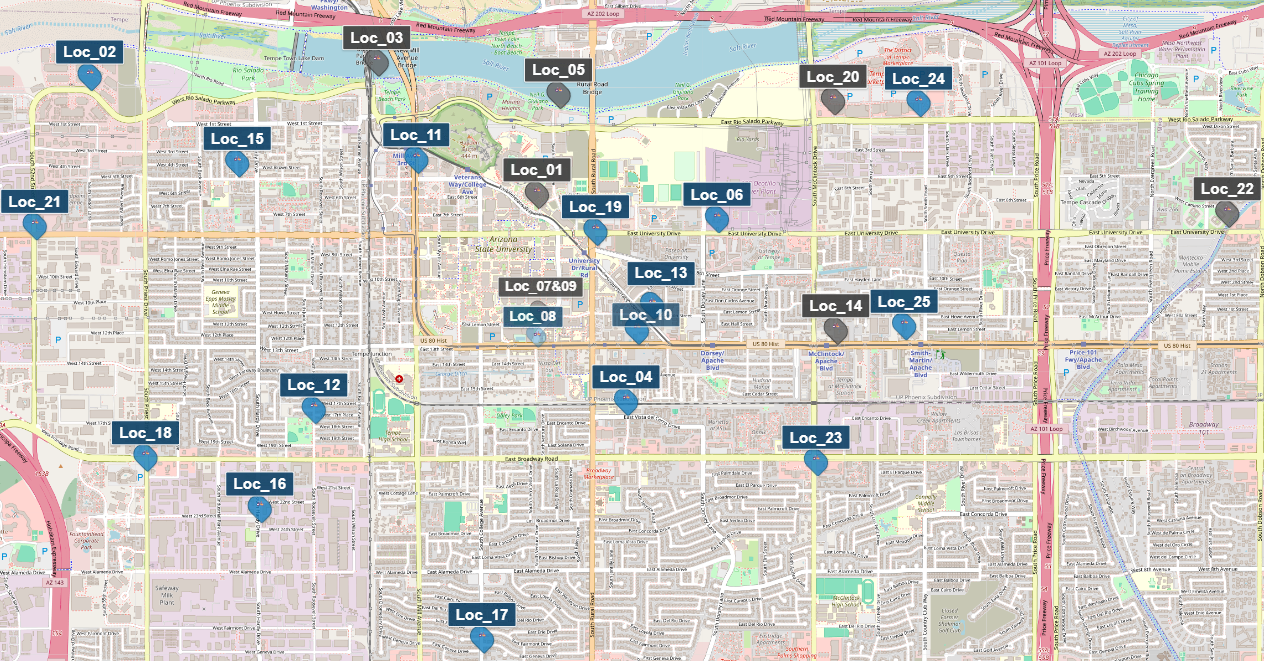}
    \caption{Spatial distribution of the fixed-camera deployment
    locations across Tempe, AZ. Locations cover campus streets,
    residential neighborhoods, urban arterials, and recreational
    corridors. Blue markers show the 19 analyzed deployments at 17
    physical locations; gray markers show the 10 excluded deployments:
    crossing-dominated fields of view (Loc\_03, 05, 05-2, 07, 07-2,
    09), unreliable annotation (Loc\_01, 20, 22), and no usable
    captures (Loc\_14). The adjacent Loc\_07 and Loc\_09 share one
    marker, and repositioned deployments (Loc\_04-2, 05-2, 07-2,
    19-2) share the marker of their base location. Basemap:
    OpenStreetMap contributors.}
    \label{fig:map}
\end{figure}

The data come from a fixed-camera project covering 25 urban locations in Tempe, Arizona (Figure~\ref{fig:map}) \citep{Lemarchand2025}. The cameras are motion-triggered wildlife cameras installed at fixed positions. When motion is detected, a camera captures two consecutive still images approximately two seconds apart, rather than a continuous video stream. Data were collected from January to April 2023, covering daytime and nighttime conditions. At four locations the camera was repositioned during the collection period; each such pair records the same physical location from a different view and period. All study locations sit on the urban street network; freeways prohibit bicycle traffic and fall outside the scope of this study under the FHWA functional classification \citep{FHWA2023}.

Not every deployment supports the direction analysis. At six deployments (Loc\_03, Loc\_05, Loc\_05-2, Loc\_07, Loc\_07-2, and Loc\_09) the camera field of view is dominated by a street crossing. Crossing movements run perpendicular to the near-side flow, so with or against that flow is undefined for them, and these deployments were excluded. Three further deployments (Loc\_01, Loc\_20, and Loc\_22) could not be annotated reliably, and one (Loc\_14) produced no usable captures. The analysis set is therefore 19 camera deployments at 17 physical locations. Within this set, Loc\_21 is also crossing-heavy (9 of its 13 movements cross the street) and Loc\_06 is a shared-use area; both are retained for the usage analysis and read with caution on direction. All observations were made in public street space; the analysis operates on anonymous event-level records carrying only timestamps, facility labels, and travel direction.

\begin{figure}[!htbp]
    \centering
    \includegraphics[width=0.6\linewidth]{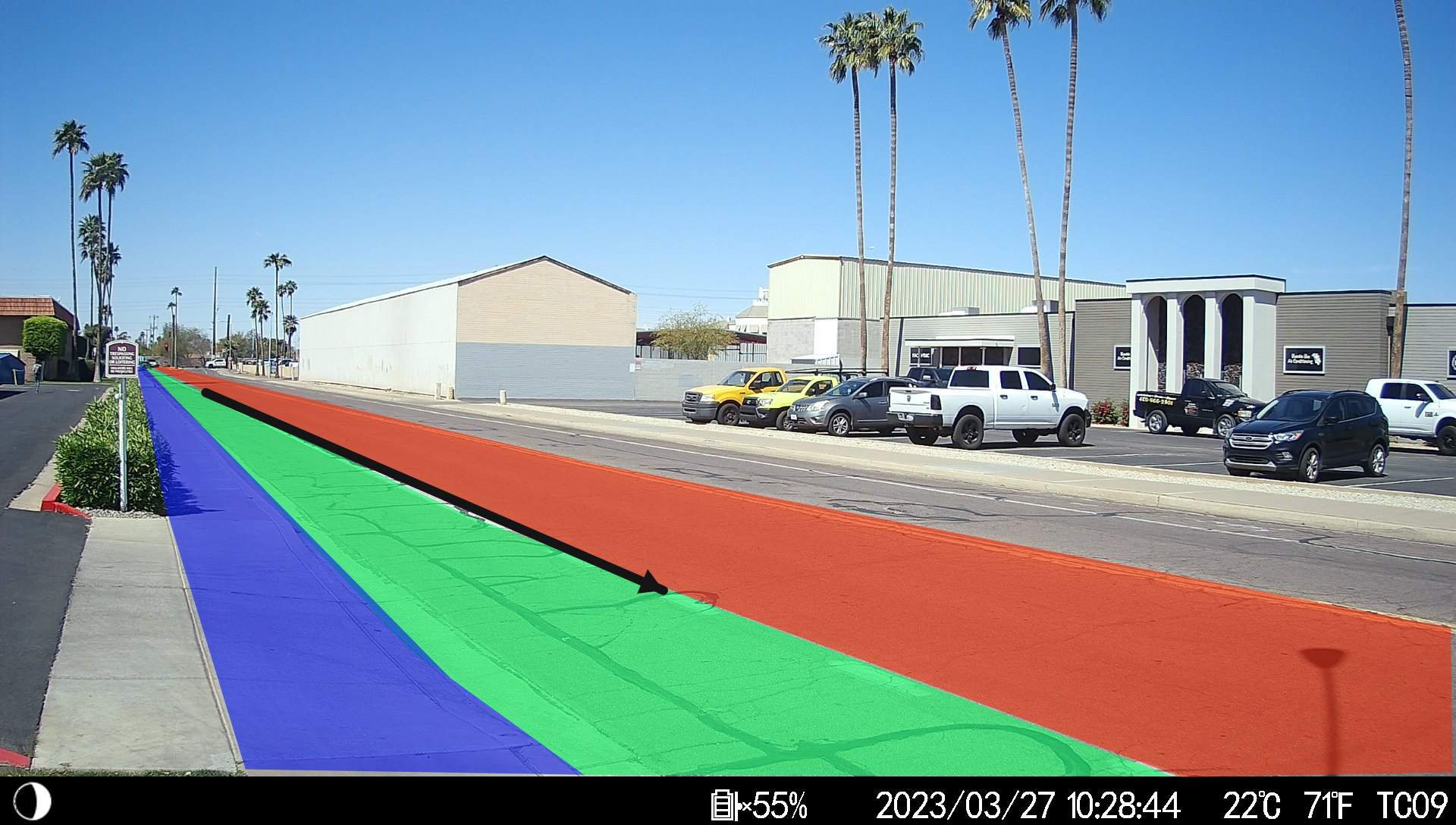}
    \caption{Example of facility annotation at a representative
    location. The blue region marks the sidewalk, the green region
    marks the bike lane, and the red region marks the travel lane. The
    black arrow indicates the reference flow direction used for
    wrong-way riding assessment.}
    \label{fig:roi}
\end{figure}

Each deployment was annotated once on its fixed camera view. For each camera view, we annotated the visible facility types as polygons in the image: the sidewalk, the bike lane, and the travel lane, matching whatever exists at that location. We also marked the reference flow direction of the near-side lane with an arrow. The direction of each arrow was determined from motor vehicle movement observed across multiple captures, together with roadway markings where visible, rather than assumed from map data. Figure~\ref{fig:roi} shows the annotation at a representative location. All annotations were re-checked against street imagery during the study; one location initially missing its bike lane polygons (the lanes are not painted green there) was corrected, and all others were verified.

\begin{figure}[!htbp]
    \centering
    \includegraphics[width=0.82\linewidth]{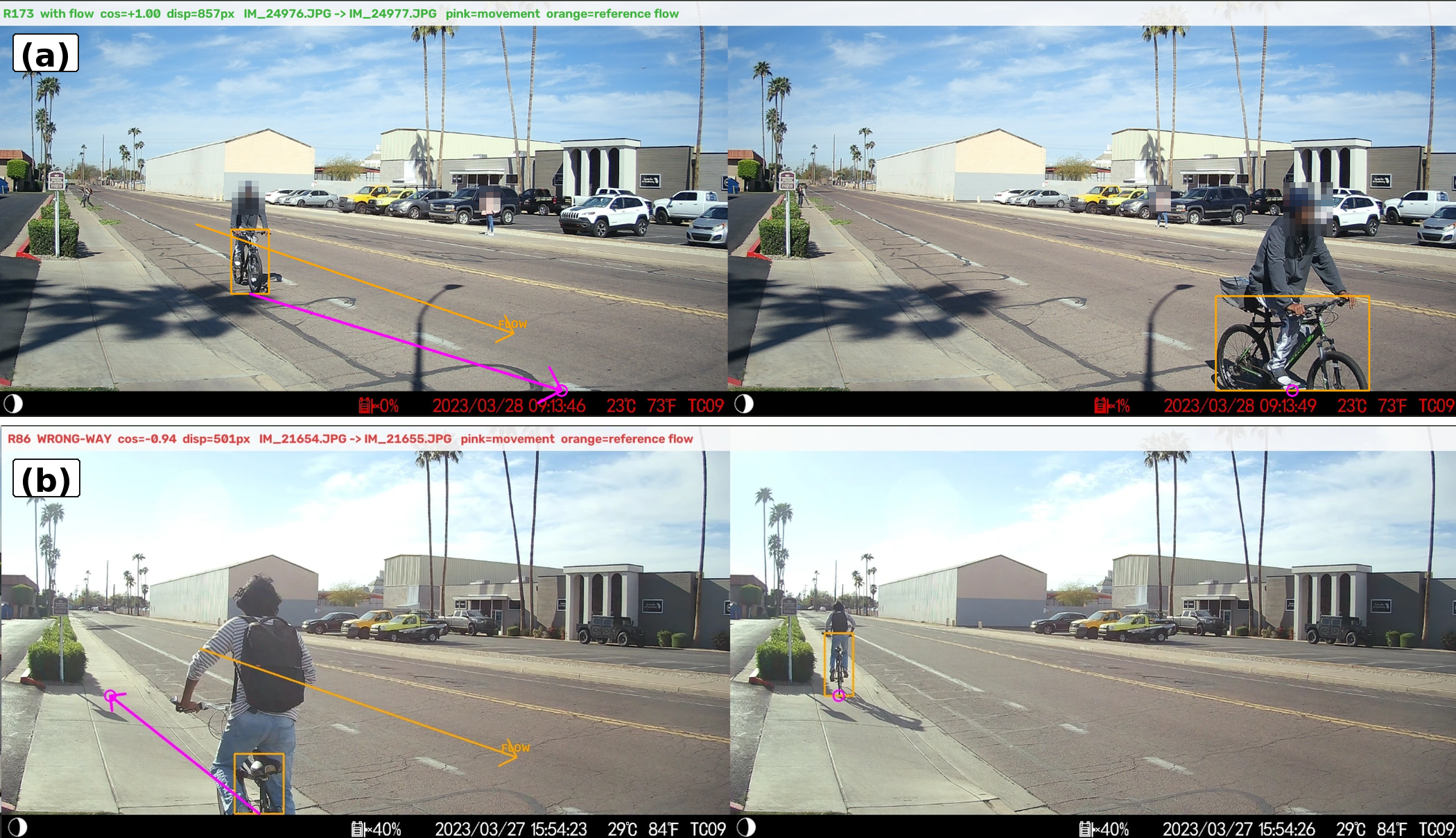}
    \caption{Verification images for two cycling events at the same
    deployment, exactly as exported by the pipeline for review.
    (a)~An event labeled with-flow: the displacement between the two
    captures (pink arrow) aligns with the reference flow direction
    (orange arrow; cosine $+1.00$). (b)~An event labeled wrong-way: the
    displacement opposes the reference flow (cosine $-0.94$). Every
    wrong-way call in this study was reviewed against such an image.
    Faces are blurred for publication.}
    \label{fig:event}
\end{figure}

The basic unit of the dataset is the cycling event: one cyclist passing through one camera view. Figure~\ref{fig:event} shows two events as recorded by the camera's double-shot, one with the flow and one against it. Each event record carries the location, the timestamps, the facility the cyclist mainly used, and the travel direction where it could be judged.

Two datasets are built from the same captures. First, a complete manual count: the first author reviewed every capture at the 19 deployments and recorded each cycling event with its facility and travel direction. This review identified 1,963 cycling events; 20 of them were roadway crossings perpendicular to the near-side flow (3 at Loc\_04, 8 at Loc\_15, and 9 at Loc\_21) and were excluded under the crossing treatment above, leaving 1,943 events. Because a human can judge heading from a single image, the manual count assigns a direction to every event. It is the ground truth for all facility and direction statistics in this paper. Second, an automated pipeline processes the same captures without human input; its outputs are validated against the manual count in the Results. The unit of analysis is the camera deployment: event-level records are aggregated into deployment-level metrics, since individual events are subject to random variation while aggregates reflect stable usage patterns.

\section{Methodology}

\begin{figure}[!htbp]
    \centering
    \includegraphics[width=\linewidth]{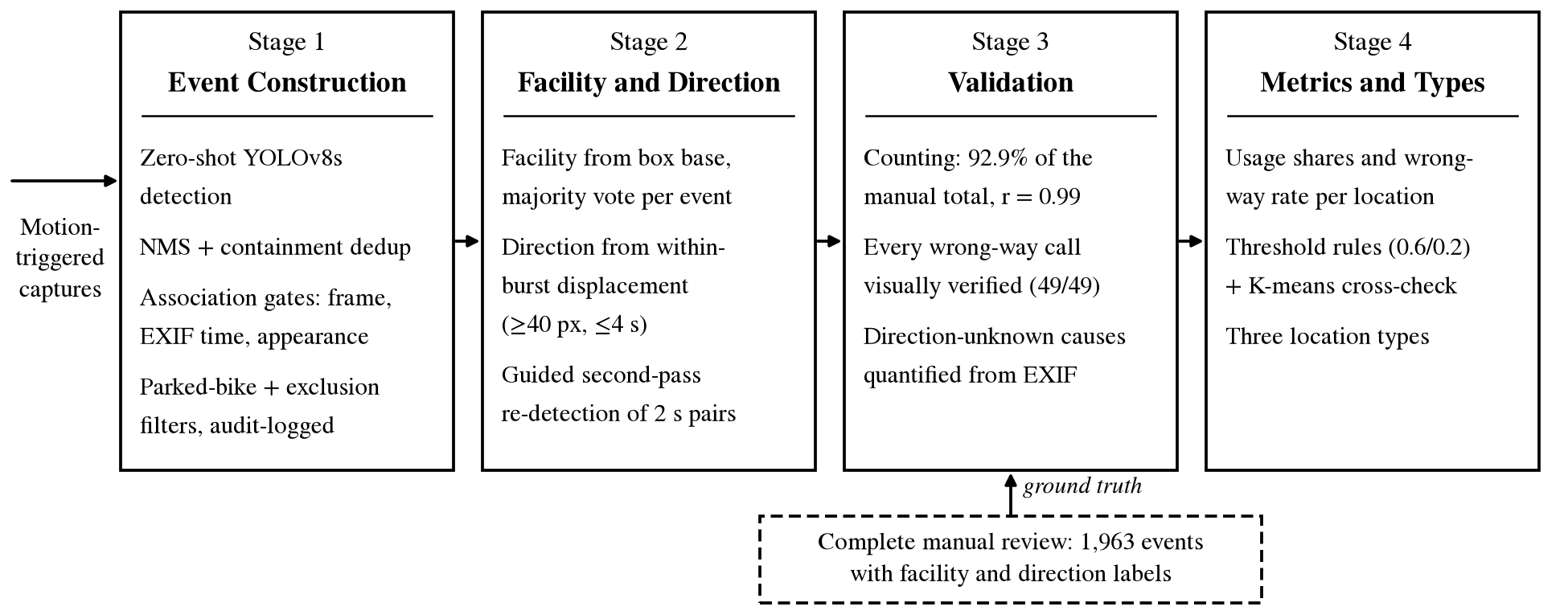}
    \caption{The four-stage workflow. Motion-triggered captures are
    grouped into cycling events, each event receives facility and
    direction labels, the automated outputs are validated against a
    complete manual count, and deployment-level metrics classify each
    deployment into one of three location types.}
    \label{fig:pipeline}
\end{figure}

This section presents the four-stage workflow (Figure~\ref{fig:pipeline}). Stage 1 constructs events from the raw captures. Stage 2 infers facility use and travel direction for each event. Stage 3 validates the automated outputs against the manual count. Stage 4 aggregates events into deployment-level metrics and classifies deployments into types.

\subsection{Stage 1: Event Construction}

Cyclist observations come from a pretrained YOLOv8s detector run on each capture at a confidence threshold of 0.10 and an inference size of 1,280 pixels; the low threshold and high resolution preserve recall on small and nighttime targets. The detector is used zero-shot: an off-the-shelf model with no training or fine-tuning on this study's images. Duplicate boxes on one bicycle are removed by non-maximum suppression plus a containment test that catches nested boxes.

Observations are then associated into events. Two observations join the same event only if they pass three gates. The frame gate requires their capture positions in the sequence to differ by at most 3. The time gate requires their EXIF capture times to differ by at most 30 seconds; this matters because the cameras are motion-triggered, so consecutive image numbers can be minutes apart and belong to different cyclists. The appearance gate requires their color fingerprints to agree: each observation carries a hue-saturation histogram plus a brightness histogram, and the two are compared separately, so that a white-shirted cyclist is never chained to a dark-jacketed one even when their positions align. Two further filters remove non-cyclists. A parked bicycle triggers a detection at the same spot every time a passer-by trips the shutter; positions hit repeatedly over spans of five minutes or more, in most captures of their time window, are removed as stationary objects. Recurring false positives at fixed spots (tree foliage detected as a bicycle at night, at one location) are removed through human-verified exclusion zones. Every removal is logged to an audit file, so each filtering decision can be re-examined.

\subsection{Stage 2: Facility Use and Direction Inference}

Each observation is assigned to a facility using the lower portion of its bounding box as a proxy for ground contact, so a cyclist is assigned to the facility their wheels are on. Overlapping polygons resolve by priority (crosswalk, then bike lane, then sidewalk, then travel lane). At the event level, the facility used in the largest number of observations becomes the event's dominant facility; this matches the definition a human counter uses.

Travel direction is inferred from displacement, and only where the displacement can be trusted. The camera's two-second double-shot provides the natural measurement pair: direction is computed only when an event has two observations within a four-second window whose displacement is at least 40 pixels. The displacement is compared with the reference flow by its cosine. Movement mostly perpendicular to the flow (absolute cosine below 0.5) is labeled crossing, for which wrong-way is undefined; otherwise the event is labeled with or against the flow. Chains spanning more than one trigger are kept for counting but contribute no direction, because visual review showed such chains frequently merge two different cyclists. When the detector finds a cyclist in only one image of a pair, a guided second pass re-runs detection on the other image at a lower threshold and accepts a recovery only if it matches the cyclist's position range and appearance fingerprint. Events that fail these requirements are labeled direction-unknown by the pipeline; they remain in the facility analysis. The manual count is not affected by any of this: a human assigns direction to every event from the image itself.

The analysis covers the near side of the street. The reference flow is defined for the lane closest to the camera, because far-side cyclists are frequently occluded by traffic and a two-way street has no single reference direction. Sidewalks carry two-way pedestrian traffic and bear no directional markings, so wrong-way rates on sidewalks measure directional mixing relative to the near-side flow; the interpretation below rests on this distinction.

\subsection{Stage 3: Validation Against the Manual Count}

Automated outputs are evaluated against the manual ground truth at the level each output supports: counting and facility use at the deployment level, and direction through a pooled comparison plus manual review of every automated wrong-way call. Because the dataset carries no ground-truth bounding boxes, standard detector metrics such as mAP or a confusion matrix cannot be computed, and because the manual count records totals rather than a matched event log, event-level precision and recall are equally unavailable. Counting is compared per deployment and in aggregate; because the pipeline does not exclude crossing riders, its count is compared with the full manual review count of 1,963 events, so both sides of the comparison cover the same population. Direction is validated in three ways. First, the pooled wrong-way rate of the pipeline's direction-known subset is compared with the manual rate. Second, every wrong-way call the pipeline makes is exported as a side-by-side verification image (first and last capture, with the displacement and reference arrows drawn) and visually reviewed; Figure~\ref{fig:event} shows two such images. Direction failure modes found in review (chains merging two cyclists across triggers, detection boxes jumping between cyclists in busy scenes, one animal false positive) were fixed by the gates of Stages 1 and 2, and the review was repeated on the final outputs. Third, the reasons for direction-unknown events are quantified from the EXIF record, so coverage is a measured property rather than an unexplained gap. Results appear in the next section.

\subsection{Stage 4: Location Metrics and Classification}

Location metrics are computed from the manual ground truth, which covers every event. Each deployment gets four numbers: the shares of events whose dominant facility is the sidewalk, the bike lane, and the travel lane (these are exclusive and sum to one), and the wrong-way rate over all events. These three facilities are the bicycle facility types that occur across the analyzed locations, drawn from the set that national design guidance defines for urban streets \citep{AASHTO2024}; a location can be dominated by any one of them, or by none. A deployment is classified as bike-lane dominant, sidewalk dominant, or travel-lane dominant if its highest share is at least 0.6 and exceeds the second-highest by at least 0.2; otherwise it is mixed. In this dataset three of the four possible types occur; no analyzed location is travel-lane dominant, since locations without separated facilities in this deployment are the crossing-dominated ones excluded from the direction analysis. The assignments are stable to reasonable variations in the thresholds: moving the dominance threshold to 0.55 or 0.65 and the gap threshold to 0.15 or 0.25 changes no assignment, and a K-means clustering ($k = 3$) of the same usage shares recovers the identical grouping at all 19 deployments (adjusted Rand index 1.0). Silhouette scores do not single out $k = 3$ (0.67, 0.61, 0.68, and 0.69 for $k = 2$ through $5$); that $k$ matches the granularity of the three-type taxonomy, and the exact agreement at it is the check reported here. The wrong-way rate is kept out of the classification, so it remains an independent measure of each location.
\section{Results}

Table~\ref{tab:locations} lists all 19 deployments with their events, usage shares, and wrong-way riding rates from the manual ground truth, together with the pipeline's event count; every number discussed below can be traced to a row of this table.

\begin{table}[htbp]
\caption{Cycling events, facility usage shares, and wrong-way riding
rate for the 19 analyzed deployments, grouped by location type. Events
and all shares come from the complete manual count; Pipeline is the
automated pipeline's event count at the same deployment; the pipeline does not exclude crossing riders, so its count validation in the text uses the full manual review count (1,963 events, adding the crossings: 3 at Loc\_04, 8 at Loc\_15, 9 at Loc\_21). SW, BL, RD =
shares of events whose dominant facility is the sidewalk, bike lane,
and travel lane; the three shares are exclusive and sum to 1.00 up to
rounding. WW is
the wrong-way riding rate over all events, reported as rate ($n/N$),
with the Wilson 95\% interval. Rates at deployments with fewer than 10
events (Loc\_02, 21, 25) carry wide intervals and should be read with
caution.}
\label{tab:locations}
\centering
\small
\renewcommand{\arraystretch}{0.97}
\setlength{\tabcolsep}{4.5pt}
\begin{tabular}{lrrrrrlc}
\toprule
Location & Events & Pipeline & SW & BL & RD & WW rate ($n/N$) & 95\% CI \\
\midrule
\multicolumn{8}{l}{\textit{Bike-lane dominant (9 deployments)}} \\
Loc\_17 & 420 & 458 & 0.00 & 0.99 & 0.01 & 0.01 (6/420) & 0.01--0.03 \\
Loc\_12 & 101 & 89 & 0.01 & 0.97 & 0.02 & 0.08 (8/101) & 0.04--0.15 \\
Loc\_15 & 81 & 91 & 0.30 & 0.70 & 0.00 & 0.12 (10/81) & 0.07--0.21 \\
Loc\_19 & 75 & 55 & 0.05 & 0.83 & 0.12 & 0.16 (12/75) & 0.09--0.26 \\
Loc\_13 & 72 & 62 & 0.25 & 0.75 & 0.00 & 0.21 (15/72) & 0.13--0.32 \\
Loc\_16 & 67 & 61 & 0.27 & 0.73 & 0.00 & 0.16 (11/67) & 0.09--0.27 \\
Loc\_11 & 67 & 58 & 0.00 & 0.99 & 0.01 & 0.01 (1/67) & 0.00--0.08 \\
Loc\_19-2 & 39 & 31 & 0.18 & 0.79 & 0.03 & 0.23 (9/39) & 0.13--0.38 \\
Loc\_02 & 2 & 4 & 0.00 & 1.00 & 0.00 & 0.00 (0/2) & 0.00--0.66 \\
\midrule
\multicolumn{8}{l}{\textit{Sidewalk dominant (6 deployments)}} \\
Loc\_08 & 214 & 183 & 0.69 & 0.31 & 0.00 & 0.43 (93/214) & 0.37--0.50 \\
Loc\_10 & 160 & 143 & 0.71 & 0.29 & 0.00 & 0.37 (59/160) & 0.30--0.45 \\
Loc\_06 & 100 & 74 & 0.81 & 0.19 & 0.00 & 0.38 (38/100) & 0.29--0.48 \\
Loc\_23 & 93 & 91 & 0.70 & 0.30 & 0.00 & 0.34 (32/93) & 0.26--0.45 \\
Loc\_18 & 23 & 17 & 0.78 & 0.22 & 0.00 & 0.26 (6/23) & 0.13--0.46 \\
Loc\_21 & 4 & 10 & 1.00 & 0.00 & 0.00 & 0.25 (1/4) & 0.05--0.70 \\
\midrule
\multicolumn{8}{l}{\textit{Mixed (4 deployments)}} \\
Loc\_04 & 232 & 220 & 0.52 & 0.46 & 0.02 & 0.27 (62/232) & 0.21--0.33 \\
Loc\_04-2 & 162 & 146 & 0.54 & 0.46 & 0.01 & 0.28 (45/162) & 0.21--0.35 \\
Loc\_24 & 25 & 27 & 0.52 & 0.48 & 0.00 & 0.28 (7/25) & 0.14--0.48 \\
Loc\_25 & 6 & 3 & 0.50 & 0.50 & 0.00 & 0.50 (3/6) & 0.19--0.81 \\
\midrule
Total & 1,943 & 1,823 & & & & 418/1,943 & \\
\bottomrule
\end{tabular}
\end{table}

\subsection{Pipeline Validation}

\begin{figure}[!htbp]
    \centering
    \includegraphics[width=\linewidth]{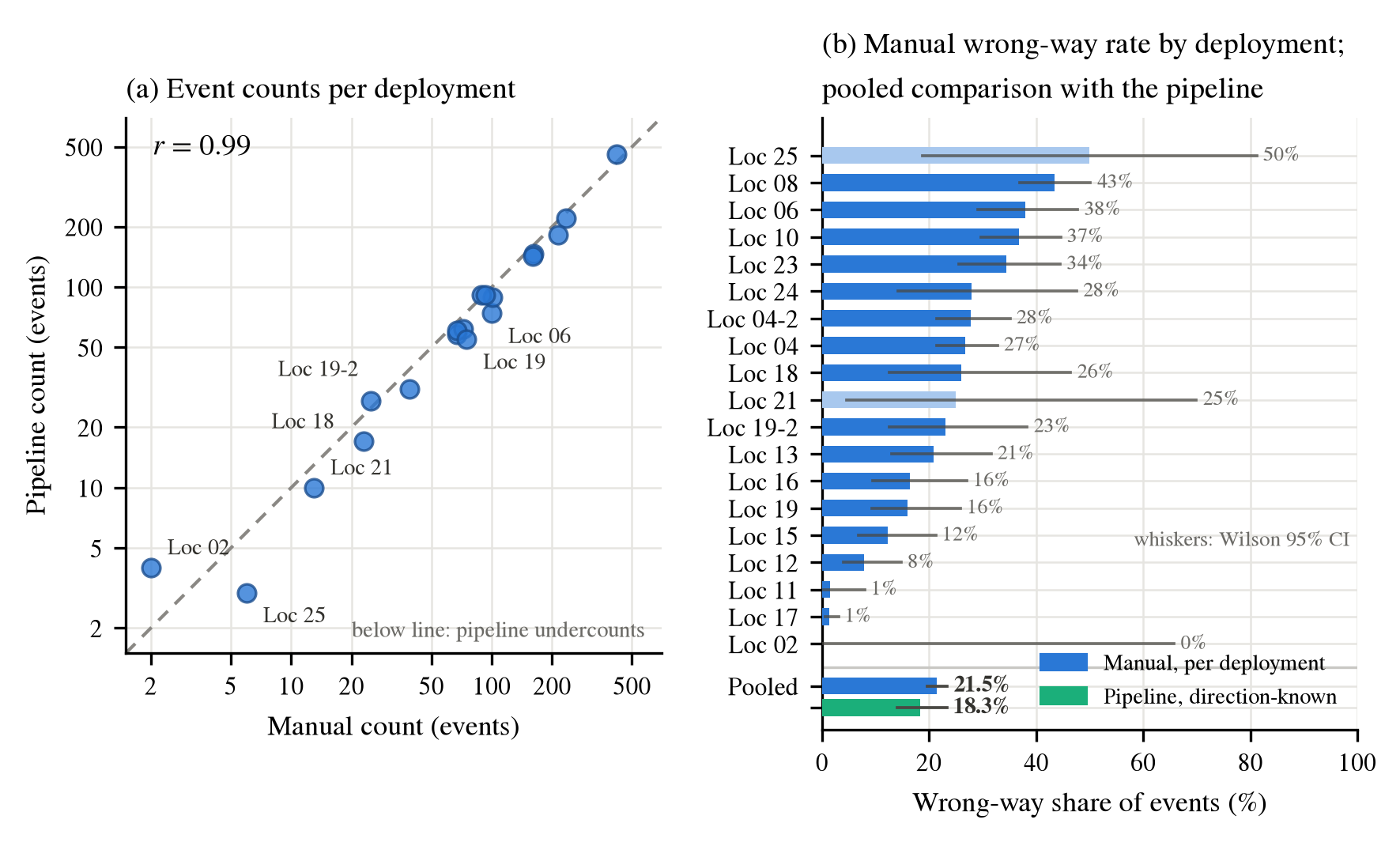}
  \caption{Validation of the automated pipeline against the manual
benchmark. (a)~Pipeline event count versus manual count per
deployment (log scales, Pearson $r$ = 0.99); manual counts here
include crossing riders, which the pipeline also counts, so both axes
cover the same population; the dashed line marks perfect agreement,
and deployments deviating by more than 20 percent are labeled. (b)~Manual wrong-way rate at each of the 19 deployments
(Wilson 95\% intervals), sorted by rate; lightened bars mark the
three deployments with fewer than 10 events (Loc\_02, 21, 25); the
pooled pair beneath
compares the manual rate (21.5 percent, 418 of 1,943 events) with the
pipeline's direction-known subset (18.3 percent, 49 of 268), whose
interval includes the manual rate. Per-deployment pipeline rates are
not shown because at 18 of the 19 deployments they rest on fewer than
36 direction calls each, too few for stable rates; the pooled
comparison checks aggregate agreement rather than event-level
direction accuracy, and every one of the 49 wrong-way calls was
individually confirmed by visual review.}
    \label{fig:validation}
\end{figure}

The automated pipeline constructs 1,823 events. Because it does not exclude crossing riders, its counting is compared with the full manual review count of 1,963 events: the count ratio is 92.9 percent, and its per-deployment counts track the manual counts closely (Pearson $r$ = 0.99; Figure~\ref{fig:validation}a). Setting aside Loc\_25 (3 pipeline events against 6 manual), the count ratio is lowest at three deployments with substantial nighttime traffic (0.73 to 0.74 at Loc\_06, Loc\_18, and Loc\_19) where motion blur and occlusion cost detections, and at the small Loc\_21 (10 against 13); it is 0.85 or higher at 13 of the 19 deployments. Where the pipeline exceeds the manual count (four deployments, 44 surplus events, 2.4 percent of the pipeline total), review of flagged cases identified track fragmentation, one cyclist split into two events, as the main mechanism rather than phantom detections; it drives the largest surplus (Loc\_17, 38 events). Static false positives are removed separately and logged (56 detections at a single foliage spot at one location). The facility assignment is validated the same way: running the classification rules of Stage 4 on the pipeline's own usage shares reproduces the manual type at 17 of 19 deployments, and the pipeline's facility shares track the manual shares closely at the 16 deployments with at least 10 events ($r$ = 0.98, mean absolute error 0.05 per share). Both disagreements sit at the smallest deployments (Loc\_21, 4 manual events, where the pipeline also counts crossing riders; Loc\_25, 6).

On direction, the pipeline is deliberately conservative: it assigns a direction only from a verified two-second pair, which covers 14.7 percent of its events. Within that subset, its pooled wrong-way rate is 18.3 percent (49 of 268), against 21.5 percent (418 of 1,943) in the full manual count (Figure~\ref{fig:validation}b). Every one of the 49 wrong-way calls was visually reviewed against its verification image and confirmed. This subset is not a random sample of events: it requires a clean two-second pair, so it favors unobstructed riders with clear movement, and the pooled comparison checks aggregate agreement rather than event-level direction accuracy. The 85.3 percent of events without an automated direction decompose as follows from the EXIF record: 48 percent have no second capture within four seconds on disk, so no displacement exists to measure; 26 percent have the pair, but the detector missed the cyclist in one image (night blur, occlusion, or the cyclist leaving the frame); 25 percent were associated across captures but their direction was withheld by the conservative gates; 1 percent moved too little to judge. Direction coverage is therefore bounded mainly by the archive itself, which is why the manual count, where a human judges heading from a single image, serves as the direction ground truth throughout.

\subsection{Usage Patterns Across Locations}

Under the threshold rules, the 19 deployments fall into three types (Table~\ref{tab:type_summary}). Bike-lane dominant is the largest group: 9 deployments with a mean bike lane share of 0.86. Sidewalk dominant covers 6 deployments with a mean sidewalk share of 0.78. Mixed covers 4 deployments where sidewalk and bike lane use coexist at nearly equal shares. Travel lane riding is rare throughout the sample: 22 of 1,943 events (1.1 percent) have the travel lane as their dominant facility, and no deployment's travel lane share exceeds 0.12, which is why the fourth possible type does not occur. The four repeated deployments offer an internal consistency check: in both pairs (Loc\_04 with Loc\_04-2, and Loc\_19 with Loc\_19-2), the two views of the same physical location receive the same type and closely aligned usage profiles.

\begin{table}[!htbp]
\caption{Summary statistics for each location type. SW, BL, RD = mean
shares of events whose dominant facility is the sidewalk, bike lane,
and travel lane; WW = mean wrong-way riding rate over all events;
means are unweighted across deployments. Boldface marks the share
that defines each type.}
\label{tab:type_summary}
\centering
\small
\begin{tabular}{lrrrrrr}
\toprule
Type & Deployments & Events & SW & BL & RD & WW \\
\midrule
Bike-lane dominant & 9 & 924 & 0.12 & \textbf{0.86} & 0.02 & 0.11 \\
Sidewalk dominant  & 6 & 594 & \textbf{0.78} & 0.22 & 0.00 & 0.34 \\
Mixed              & 4 & 425 & 0.52 & 0.47 & 0.01 & 0.33 \\
\midrule
Total              & 19 & 1,943 & & & & \\
\bottomrule
\end{tabular}
\end{table}

The types give the first hypothesis qualified support. Facility availability constrains where cycling can occur: where a bike lane dominates the profile, cyclists are concentrated in it. But presence alone does not determine dominant use. A bike lane is present at every analyzed deployment except Loc\_21, including every sidewalk-dominant one, and cyclists ride the sidewalk anyway; at Loc\_08, 69 percent of 214 events kept to the sidewalk beside an available bike lane. Configuration is a necessary but not sufficient condition for how a location is used.

\subsection{Facility Use and Travel Direction}

The second hypothesis concerns the link between bike lane use and wrong-way riding, and here the manual ground truth allows a direct answer at the event level. Across all 1,943 events, cyclists whose dominant facility was the bike lane rode against the reference flow in 4.6 percent of cases (55 of 1,195). Cyclists on sidewalks did so in 49.6 percent (360 of 726), close to the coin-flip share expected on a facility that marks no direction. Travel lane events are too few to interpret (3 of 22). Wrong-way riding in this dataset is not a general behavior spread across the network; it is concentrated where riding happens on sidewalks. The link is not purely composition either. Among bike-lane-dominant deployments the wrong-way rate within bike lane events varies widely: 1.0 percent at Loc\_17 (4 of 417) against 19.4 percent at Loc\_19 (12 of 62) and 16.1 percent at its repeat view Loc\_19-2 (5 of 31). Deployments differ in behavior beyond their facility mix, and the two views of the same physical location agree.

\begin{figure}[!htbp]
    \centering
    \includegraphics[width=0.62\linewidth]{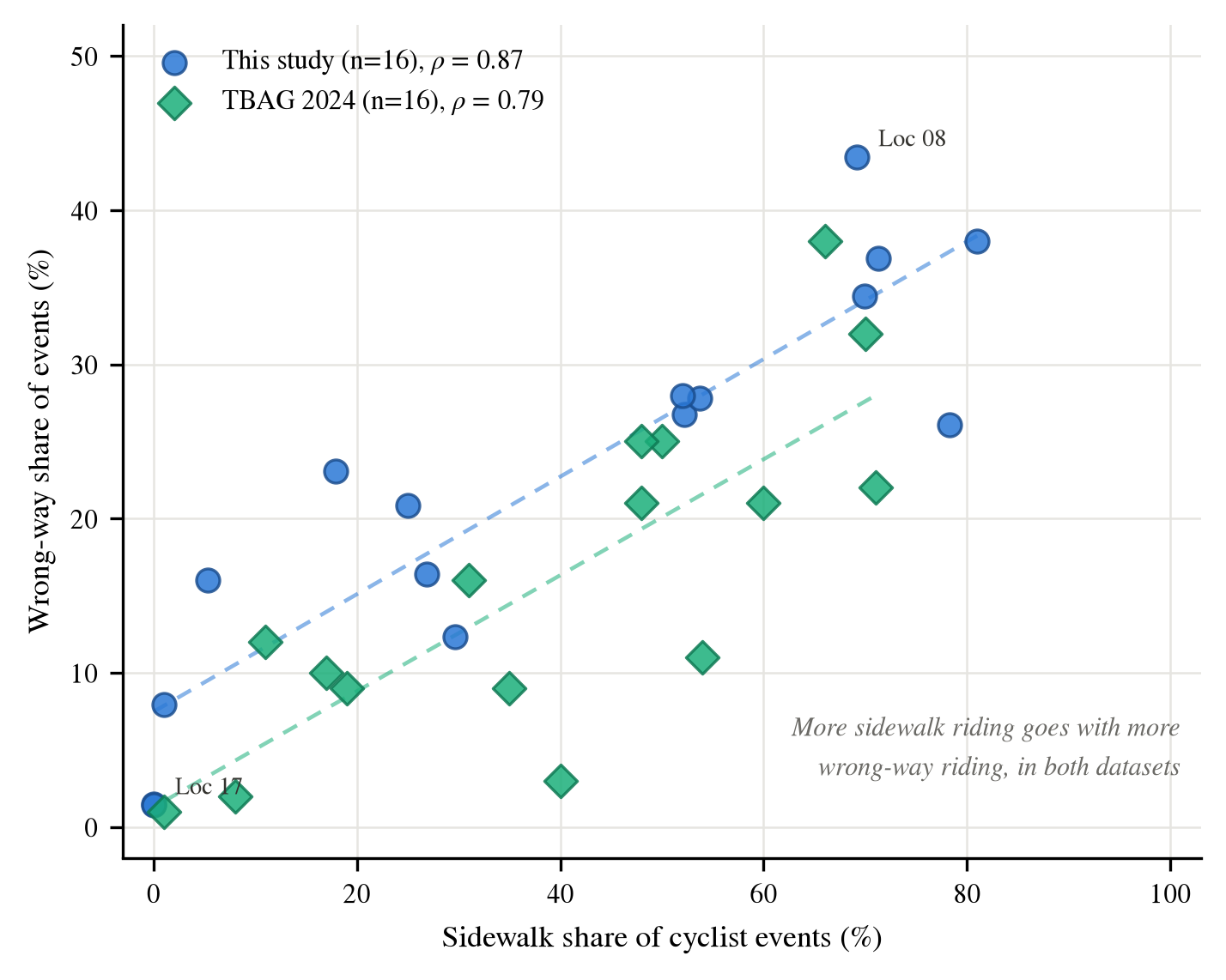}
    \caption{Sidewalk share of cycling events versus wrong-way riding
    rate. Circles: the 16 deployments of this study with meaningful
    volume, from the manual ground truth. Diamonds: 16 intersections
    from the independent Tempe Bicycle Action Group 2024 count.
    Dashed lines are linear fits to each dataset, drawn as visual
    guides. Both datasets show the same strong positive association.}
    \label{fig:scatter}
\end{figure}

The location level shows the same structure. Figure~\ref{fig:scatter} plots the sidewalk share of each deployment against its wrong-way rate, for the 16 deployments with meaningful volume (excluding the three with fewer than 10 events, per Table~\ref{tab:locations}). The association is strong (Spearman $\rho = 0.87$, $p < 0.001$); equivalently, bike lane share and wrong-way rate correlate at $\rho = -0.87$. The second hypothesis is supported: locations where cycling stays in the bike lane have almost no wrong-way riding (Loc\_17: 6 of 420 events; Loc\_11: 1 of 67), while the sidewalk-dominant deployments in this subset run between 0.26 and 0.43.

An independent dataset shows the same structure. The Tempe Bicycle Action Group's 2024 volunteer count recorded sidewalk share and wrong-way share at 16 intersections in the same city \citep{TBAG2024}. Their published overall wrong-way share is 14.1 percent against 21.5 percent here, their per-intersection range (1 to 38 percent) matches ours (1 to 43 percent across the 16 deployments with at least 10 events), and the correlation between sidewalk share and wrong-way rate in their data is $\rho = 0.79$ ($p < 0.001$), computed from the per-intersection counts in Table G1 of their report. The two instruments differ in scope. The volunteers record all four approaches of an intersection during peak windows and count wrong-way riding as cycling against traffic on any facility, while this study covers the near-side flow around the clock; and the volunteer count publishes no facility-by-direction cross table, so it can confirm the location-level association but not the event-level split between bike lane and sidewalk. In both datasets the association also partly reflects composition, since wrong-way movement comes mostly from sidewalk riders; the within-facility variation reported above is what goes beyond composition. The count has run since 2011, and its history brackets this study's levels: across the 2011 to 2018 counts its overall wrong-way share ranged from 16.6 to 21.9 percent (14.1 in 2024) and its sidewalk share from 27.7 to 46.2 percent (34.5 in 2024), and the 2023 observations here sit inside both ranges, at 21.5 percent wrong-way and 37.4 percent sidewalk. Figure~\ref{fig:scatter} overlays both datasets: two instruments with different observers, sampling windows, and units of observation agree on levels as well as on the direction of the association.

Taken together, facility presence is strongly tied to where cyclists ride, and where cyclists ride is strongly tied to their direction. A location's directional problem is usually a usage problem seen from another angle: activity that has shifted onto the sidewalk, where traffic runs both ways and directional order dissolves. Evaluating a facility by volume alone would miss this entirely.

\section{Discussion}

The three location types reflect how cyclists arrange themselves under different facility configurations. Where a bike lane dominates, cyclists are concentrated in it and almost all ride with the flow. Where activity has shifted to the sidewalk, direction mixes. The mixed type sits in between on both measures. The consistency of these types across heterogeneous locations, and across repeated deployments at the same location, suggests that where cyclists ride is largely shaped by what the street offers.

The direction finding reframes the wrong-way question. In this dataset, wrong-way riding is not primarily a story of cyclists violating the direction of bike lanes; bike lane events ran against the flow in only 4.6 percent of cases. It is a story of cyclists riding on sidewalks, where traffic runs both ways and half of all movement runs against the near-side flow almost by construction. This mixing is not sanctioned by rule: Tempe permits adult sidewalk riding only under specified conditions \citep{TempeCode2019}, and city guidance advises sidewalk riders to travel with the flow of adjacent traffic \citep{TempeBikeSafety}, yet half of the sidewalk events ran against it. Rider age cannot be determined from these captures, and minors riding with a guardian's consent are exempt from the sidewalk restriction, so we report directional mixing as measured behavior rather than as a count of violations. The distinction matters for safety interpretation. A cyclist entering an intersection from a sidewalk against the adjacent traffic direction is a well-documented crash configuration \citep{Hunter1996}, so a high sidewalk share marks elevated directional conflict, and the behavior follows the facility's design rather than its regulation. Why cyclists choose the sidewalk despite an available bike lane is beyond what these cameras record: comfort, perceived safety, or network connectivity are plausible reasons, and separating them is work for survey-based studies. The sidewalk pattern is also not the only one. At Loc\_19 and its repeat view Loc\_19-2, two camera positions at the same physical location, wrong-way riding inside the bike lane runs several times the pooled bike lane rate, while the sidewalk share there is only 0.05. A screen built on sidewalk share alone would not flag it. Directional mixing on sidewalks is the dominant pattern in this dataset, but a bike lane can carry a directional problem of its own.

For practice, the results carry three messages. First, facility configuration is strongly tied to where cyclists ride, consistent with dedicated bike lanes serving their intended role. Second, a facility evaluation should ask where the activity actually is: at every sidewalk-dominant location a bike lane exists, and the directional conflict rides with the sidewalk share. Third, the workflow is a low-cost screening tool. It runs on motion-triggered still cameras without site-specific model training; each location needs a one-time annotation of facility polygons and a reference direction. Its counting is validated (92.9 percent of the full manual count, $r$ = 0.99), every automated wrong-way call is backed by a reviewable verification image (Figure~\ref{fig:event}), and every filtering decision is logged to an audit file. An agency can rank locations by sidewalk share and directional mixing and send field review where the ranking points, instead of discovering such locations through crash records after the fact.

This study has limits. The data come from one city and one collection window (January to April 2023). The manual ground truth was produced by a single counter; no second-rater reliability study was performed, though every automated wrong-way call was independently checked against its verification image. Several deployments have few events, and their rates carry wide intervals reported in Table~\ref{tab:locations}. At Loc\_15, a reflective sign leaned into the right quarter of the camera view for roughly three hours of one afternoon, a window that contained 22 of the deployment's 267 detection-bearing captures. Cyclists passing through the visible field were still counted, but short sidewalk movements confined to the obscured segment would escape both counts, so Loc\_15's sidewalk share may be slightly understated. The observations cover the near-side direction of travel, so far-side activity on two-way streets is not represented. The pipeline's count ratio is lowest at night on sidewalks, so the automated sidewalk counts under-represent nighttime activity relative to the manual count. The automated direction covers 14.7 percent of events; the decomposition in the Validation subsection shows this is mainly a property of the archive (half of the events have no usable second capture), and the manual count carries the direction analysis. Crossing-dominated deployments were excluded because a one-way reference flow does not describe crossing movements; a separate treatment of crossings is future work. Two location pairs are repeated deployments at the same physical location, so their metrics are not fully independent; dropping the second view of each pair leaves the association of Figure~\ref{fig:scatter} unchanged ($\rho = 0.87$, $n = 14$).

\section{Conclusion}

This study asked two questions about cycling infrastructure: whether usage patterns follow facility configurations, and whether higher bike lane use goes with less wrong-way riding. Using motion-triggered cameras at 19 deployments in Tempe, Arizona, we manually reviewed every capture (1,963 cycling events, of which 1,943 non-crossing events form the analysis set), validated an automated pipeline against that benchmark (a count of 92.9 percent of the full manual count, $r$ = 0.99, every automated wrong-way call visually verified), and classified the locations into three types. Both questions received clear answers. Usage follows the facilities present, with presence acting as a necessary but not sufficient condition: every sidewalk-dominant location has a bike lane that cyclists bypass. And direction follows usage: wrong-way riding occurs in 4.6 percent of bike lane events but 49.6 percent of sidewalk events, which carry two-way traffic; the deployment-level association is strong ($\rho = -0.87$ between bike lane share and wrong-way rate), and an independent volunteer count in the same city reproduces the pattern.

Future work can extend this in five directions. Applying the workflow in other cities would test whether the types and the sidewalk-direction link recur. Adding facility attributes such as lane width, protection type, and continuity could tie the types to design choices. Cyclist surveys could reach the motivations behind sidewalk riding, which cameras cannot. Single-image heading estimation \citep{Garcia2021Orientation} could raise the automated direction coverage toward the manual benchmark, removing the largest remaining gap between the pipeline and its ground truth. Finally, the multi-year records of the volunteer count \citep{TBAG2024} invite a stability test of the sidewalk-direction association over time.

\section{Acknowledgments}

The authors thank the fixed-camera deployment project team at Arizona State University for collecting and sharing the multi-site observation data used in this study. The dataset collection at Arizona State University was supported by the Zimin Foundation as part of a larger project about bicycling in Tempe, Arizona. The authors also thank the Tempe Bicycle Action Group for making their annual bike count data publicly available. During manuscript preparation, the first author used a generative AI tool (Claude) to assist with analysis code development, grammatical editing, and LaTeX formatting; all analytical decisions, manual counts, and verification reviews were performed by the authors, and this use has been disclosed in accordance with TRB policy.

\section{Author Contributions}
The authors confirm contribution to the paper as follows: study conception and design: A. He, V. W.H. Wong; data collection and curation: R. Eiris, D. Salon, M. Gupta; analysis and interpretation of results: A. He, V. W.H. Wong; draft manuscript preparation: A. He. All authors reviewed the results and approved the final version of the manuscript.

\section{Declaration of Conflicting Interests}
The authors declared no potential conflicts of interest with respect to the research, authorship, and/or publication of this article.

\section{Funding}

The dataset collection at Arizona State University was supported by the Zimin Foundation. The authors received no other financial support for the research, authorship, and/or publication of this article.

\section{Data Accessibility Statement}

The data used in this study are available from the corresponding author on reasonable request.

\clearpage

\bibliographystyle{trb}
\bibliography{refs_final}

@misc{FHWA2022Funding,
  author       = {{Federal Highway Administration}},
  title        = {{FHWA} Announces New Guidance, Increased Funding for Infrastructure Program that Helps States and Communities Create Safer, More Walkable Streets},
  howpublished = {U.S. Department of Transportation},
  year         = {2022},
  url          = {https://highways.dot.gov/newsroom/fhwa-announces-new-guidance-increased-funding-infrastructure-program-helps-states-and}
}

@misc{Caltrans2024,
  author       = {{California Department of Transportation}},
  title        = {California Investing Nearly \$1 Billion in Bicycle and Pedestrian Infrastructure Over Next Four Years},
  howpublished = {Caltrans News Release},
  year         = {2024},
  url          = {https://dot.ca.gov/caltrans-near-me/district-2/d2-news/d2-news-2024-03-25}
}

@misc{SafeTREC2025,
  author       = {{SafeTREC}},
  title        = {2025 {SafeTREC} Traffic Safety Facts: Bicycle Safety},
  howpublished = {Safe Transportation Research and Education Center, University of California, Berkeley},
  year         = {2025},
  url          = {https://safetrec.berkeley.edu/2025-safetrec-traffic-safety-facts-bicycle-safety}
}

@techreport{Hunter1996,
  author      = {William W. Hunter and Jane C. Stutts and Wayne E. Pein and Chante L. Cox},
  title       = {Pedestrian and Bicycle Crash Types of the Early 1990's},
  number      = {FHWA-RD-95-163},
  institution = {Federal Highway Administration, U.S. Department of Transportation},
  address     = {Washington, D.C.},
  year        = {1996}
}

@techreport{FDOT2021,
  author      = {{Florida Department of Transportation}},
  title       = {Florida Pedestrian and Bicycle Strategic Safety Plan},
  institution = {Florida Department of Transportation, State Safety Office},
  year        = {2021},
  url         = {https://www.alerttodayflorida.com/Files/Florida%27s%20Pedestrian%20Bicycle%20Strategic%20Safety%20Plan.pdf}
}

@techreport{WSDOT2021,
  author      = {{Washington State Department of Transportation}},
  title       = {Washington State Active Transportation Plan, 2020 and Beyond},
  institution = {Washington State Department of Transportation},
  year        = {2021},
  url         = {https://wsdot.wa.gov/sites/default/files/2021-12/ATP-2020-and-Beyond.pdf}
}

@techreport{AZDOT2024,
  author      = {{Arizona Department of Transportation}},
  title       = {Arizona Active Transportation Safety Action Plan},
  institution = {Arizona Department of Transportation},
  year        = {2024},
  url         = {https://azdot.gov/sites/default/files/2024-10/atsap-final-10-15-24-noappendices.pdf}
}

@misc{FHWA2022,
  author       = {{Federal Highway Administration}},
  title        = {Traffic Monitoring Guide},
  year         = {2022},
  howpublished = {U.S. Department of Transportation},
  url          = {https://www.fhwa.dot.gov/policyinformation/tmguide/}
}

@misc{Allen2021,
  author       = {Heather Allen},
  title        = {Walking and Cycling Gaps in Transport Research: An International Overview},
  year         = {2021},
  howpublished = {Volvo Research and Educational Foundations (VREF)},
  url          = {https://www.vref.se}
}

@article{Lemarchand2025,
  author  = {Dorian Lemarchand and Deborah Salon and Hue{-}Tam Jamme and Thomas Czerniawski},
  title   = {Counting bicycles, scooters, and skateboards in Tempe, {AZ}},
  journal = {Findings},
  year    = {2025},
  doi     = {10.32866/001c.141929}
}

@article{Silva2023,
  author  = {Cat Silva and Rolf Moeckel and Kelly Clifton},
  title   = {Comparative observational assessment of cyclists' interactions on urban streets with on-street and sidewalk bike lanes},
  journal = {Transportation Research Record},
  volume  = {2677},
  number  = {3},
  pages   = {836--848},
  year    = {2023},
  doi     = {10.1177/03611981221118539}
}

@techreport{Lowry2024,
  author      = {Michael Lowry and Don MacKenzie},
  title       = {Using Computer Vision Devices to Evaluate Bicycle and Pedestrian Improvements},
  institution = {Pacific Northwest Transportation Consortium (PacTrans), University of Washington},
  year        = {2024},
  url         = {https://hdl.handle.net/1773/52662}
}

@article{NabaviNiaki2018,
  author  = {Matin S. {Nabavi Niaki} and Nicolas Saunier and Luis F. Miranda-Moreno},
  title   = {Analysing cyclist behaviour at cycling facility discontinuities using video data},
  journal = {Transactions on Transport Sciences},
  volume  = {9},
  number  = {1},
  year    = {2018},
  doi     = {10.5507/tots.2018.001}
}

@article{Gao2020,
  author  = {Yuyan Gao and David C. Schwebel and Lingling Zhang and Wangxin Xiao and Guoqing Hu},
  title   = {Unsafe bicycling behavior in Changsha, China: A video-based observational study},
  journal = {International Journal of Environmental Research and Public Health},
  volume  = {17},
  number  = {9},
  pages   = {3256},
  year    = {2020},
  doi     = {10.3390/ijerph17093256}
}

@article{Ibrahim2021,
  author  = {Mohamed R. Ibrahim and James Haworth and Nicola Christie and Tao Cheng},
  title   = {{CyclingNet}: Detecting cycling near misses from video streams in complex urban scenes with deep learning},
  journal = {IET Intelligent Transport Systems},
  volume  = {15},
  number  = {10},
  pages   = {1185--1192},
  year    = {2021},
  doi     = {10.1049/itr2.12101}
}

@article{Ibrahim2024NearMiss,
  author  = {Mohamed R. Ibrahim},
  title   = {Computer vision and statistical insights into cycling near miss dynamics},
  journal = {Scientific Reports},
  volume  = {14},
  pages   = {21151},
  year    = {2024},
  doi     = {10.1038/s41598-024-70733-8}
}

@article{Li2025,
  author  = {Mingjie Li and Ben Beck and Tharindu Rathnayake and Lingheng Meng and Zijue Chen and Akansel Cosgun and Xiaojun Chang and Dana Kuli{\'c}},
  title   = {A benchmark for cycling close-pass detection from video streams},
  journal = {Transportation Research Part C: Emerging Technologies},
  volume  = {174},
  pages   = {105112},
  year    = {2025},
  doi     = {10.1016/j.trc.2025.105112}
}

@misc{WWC2024,
  author = {Xu, Jing and Shi, Wentao and Ren, Sheng and Zhang, Lijuan and
            Yang, Weikai and Gao, Pan and Qin, Jie},
  title  = {Fast wrong-way cycling detection in {CCTV} videos: Sparse sampling is all you need},
  year   = {2024},
  url    = {https://arxiv.org/abs/2405.07293},
  note   = {arXiv preprint arXiv:2405.07293}
}

@article{Garcia2021Orientation,
  author  = {Garc{\'i}a-Venegas, Marichelo and Mercado-Ravell, Diego A. and Pinedo-S{\'a}nchez, Luis A. and Carballo-Monsivais, Carlos A.},
  title   = {On the safety of vulnerable road users by cyclist detection and tracking},
  journal = {Machine Vision and Applications},
  volume  = {32},
  number  = {5},
  pages   = {109},
  year    = {2021},
  doi     = {10.1007/s00138-021-01231-4}
}

@article{Dhakal2018,
  author  = {Nirbesh Dhakal and Christopher R. Cherry and Ziwen Ling and Mojdeh Azad},
  title   = {Using {CyclePhilly} data to assess wrong-way riding of cyclists in Philadelphia},
  journal = {Journal of Safety Research},
  volume  = {67},
  pages   = {145--153},
  year    = {2018},
  doi     = {10.1016/j.jsr.2018.10.004}
}

@techreport{AASHTO2024,
  author      = {{American Association of State Highway and Transportation Officials}},
  title       = {Guide for the Development of Bicycle Facilities, 5th ed.},
  institution = {American Association of State Highway and Transportation Officials},
  address     = {Washington, D.C.},
  year        = {2024}
}

@techreport{FHWA2023,
  author      = {{Federal Highway Administration}},
  title       = {Highway Functional Classification: Concepts, Criteria and Procedures},
  institution = {U.S. Department of Transportation},
  address     = {Washington, D.C.},
  year        = {2023},
  url         = {https://rosap.ntl.bts.gov/view/dot/72430}
}

@techreport{TBAG2024,
  author       = {{Tempe Bicycle Action Group}},
  title        = {Tempe Bike Count Report 2024},
  institution  = {Tempe Bicycle Action Group},
  address      = {Tempe, Arizona},
  year         = {2024},
  note         = {Report date May 15, 2024; principal investigator Jack Ketcham},
  url          = {https://www.biketempe.org/wp-content/uploads/2024/07/Tempe-Bike-Count-Report-2024.pdf}
}

@misc{TempeCode2019,
  author       = {{City of Tempe}},
  title        = {Tempe City Code, Section 19-212: Riding on Sidewalks or Bicycle Lanes},
  year         = {2019},
  howpublished = {Municode Library},
  url          = {https://library.municode.com/AZ/Tempe/codes/City_Code?nodeId=CH19MOVEMUTR_ARTXIXOP_S19-212RISIBILA},
  note         = {Chapter 19, Motor Vehicles and Multimodal Traffic;
                  Ordinance No. O2019.22}
}

@misc{TempeBikeSafety,
  author       = {{City of Tempe}},
  title        = {Bike and Scooter Safety},
  year         = {2026},
  howpublished = {City of Tempe Transportation and Sustainability Department},
  url          = {https://www.tempe.gov/government/transportation-and-sustainability/transportation/bicycle-pedestrian/bike-and-scooter-safety},
  note         = {Accessed July 29, 2026}
}

\end{document}